\documentclass{article}

\usepackage{arxiv}

\usepackage{cite}
\usepackage{amsmath,amssymb,amsfonts}
\usepackage{algorithmic}
\usepackage{graphicx}
\usepackage{textcomp}
\usepackage{xcolor}
\def\BibTeX{{\rm B\kern-.05em{\sc i\kern-.025em b}\kern-.08em
    T\kern-.1667em\lower.7ex\hbox{E}\kern-.125emX}}

\usepackage[utf8]{inputenc} 
\usepackage[T1]{fontenc}    
\usepackage{hyperref}       
\usepackage{url}            
\usepackage{booktabs}       
\usepackage{nicefrac}       
\usepackage{microtype}      
\usepackage{verbatim}       
\usepackage{adjustbox}      
\usepackage{multirow, tabularx, array}        
\usepackage{enumitem}

\DeclareMathOperator*{\argmax}{arg\,max}

\newcommand{\featvisimgsize}{2.4cm}
\newcommand{\featvisminisize}{5cm}

\title{Convolutional Neural Networks Do Work with Pre-Defined Filters}

\date{} 					

\author{ 
    \href{https://orcid.org/0000-0002-7039-5189}{\includegraphics[scale=0.06]{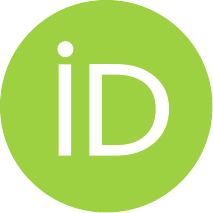}\hspace{1mm}Christoph Linse} \\
	Institute for Neuro- and Bioinformatics\\
	University of L{\"u}beck\\
	L{\"u}beck, Germany \\
	\texttt{c.linse@uni-luebeck.de} \\
	\And
	\href{https://orcid.org/000-0001-8556-2472}{\includegraphics[scale=0.06]{orcid.pdf}\hspace{1mm}Erhardt Barth} \\
	Institute for Neuro- and Bioinformatics\\
	University of L{\"u}beck\\
	L{\"u}beck, Germany \\
	\texttt{erhardt.barth@uni-luebeck.de} \\
    \And
	\href{https://orcid.org/0000-0002-4539-4475}{\includegraphics[scale=0.06]{orcid.pdf}\hspace{1mm}Thomas Martinetz} \\
	Institute for Neuro- and Bioinformatics\\
	University of L{\"u}beck\\
	L{\"u}beck, Germany \\
	\texttt{thomas.martinetz@uni-luebeck.de} \\
}

\renewcommand{\headeright}{}
\renewcommand{\undertitle}{}
\renewcommand{\shorttitle}{Convolutional Neural Networks Do Work with Pre-Defined Filters}

\hypersetup{
pdftitle={Convolutional Neural Networks Do Work with Pre-Defined Filters},
pdfauthor={Christoph Linse, Erhardt Barth, Thomas Martinetz},
pdfkeywords={convolutional neural networks, image classification, pre-defined filters, lightweight CNN},
}

\begin{document}
\maketitle

\begin{abstract}
We present a novel class of Convolutional Neural Networks called Pre-defined Filter Convolutional Neural Networks (PFCNNs), where all $n \times n$ convolution kernels with $n > 1$ are pre-defined and constant during training. It involves a special form of depthwise convolution operation called a Pre-defined Filter Module (PFM). In the channel-wise convolution part, the $1 \times n \times n$ kernels are drawn from a fixed pool of only a few (16) different pre-defined kernels. In the $1 \times 1$ convolution part linear combinations of the pre-defined filter outputs are learned. Despite this harsh restriction, complex and discriminative features are learned. These findings provide a novel perspective on the way how information is processed within deep CNNs. We discuss various properties of PFCNNs and prove their effectiveness using the popular datasets Caltech101, CIFAR10, CUB-200-2011, FGVC-Aircraft, Flowers102, and Stanford Cars. Our implementation of PFCNNs is provided on Github \url{https://github.com/Criscraft/PredefinedFilterNetworks}.
\end{abstract}


\section{Introduction}
\label{introduction}

Over the years the computer vision community has been shifting its focus from using pre-defined features towards training end-to-end systems such as Convolutional Neural Networks (CNNs), which have become state-of-the-art in many visual applications for several reasons. CNNs have empirically proven their good generalization abilities \cite{linse_large_neural_networks}, state-of-the-art performance in image recognition \cite{he_deep_2016}, especially in domains of unconstrained image data (taken in the wild) \cite{alshazly_handcrafted_2019}. The features learned by CNNs can be generic, such that they can be used for a number of different applications \cite{hertel_deep_2017}.

However, CNNs typically have millions of weights and usually operate in the over-parameterized regime, which is also called the modern interpolating regime~\cite{belkin_reconciling_2019}. Pruning experiments show that many of these weights serve no particular function and could be omitted without any loss in performance. It seems that the superiority of CNNs over traditional image recognition techniques comes with the cost of having an enormous pool of superfluous weights that makes the networks inefficient and less transparent.

In this work, we try to significantly reduce the number of trainable weights in a CNN by applying certain restrictions on the convolutional weights. We apply only a few different pre-defined kernels and do not adjust them during training to save time, computational resources, and energy.

We mold this idea into a Pre-defined Filter Module (PFM), essentially a depthwise convolution layer consisting of a channel-wise $1 \times 3 \times 3$ convolution and a subsequent $C \times 1 \times 1$ convolution that convolves over all input channels C. In the following, we will abbreviate the latter layer with $1 \times 1$ convolution. The former layer utilizes a pool of 16 different $1 \times 3 \times 3$ edge filter kernels, each of which is applied to a single input channel individually. As the pre-defined filters are frozen we only adjust the $1 \times 1$ convolution weights according to the training data. Learning in this context means finding linear combinations of the pre-defined filter outputs. Training is done end-to-end by finding linear combinations of the outputs of these pre-defined filters using gradient descent.

We construct the novel architecture PFNet18 (Pre-defined Filter Network 18) by replacing all convolution layers with $n>1$ of ResNet18 by our PFMs. This eliminates many adjustable weights and PFNet18 requires only $13\%$ of the training parameters of a ResNet18. Please note, that we use the PFMs in the entire network, not only in the first layer. We give empirical evidence that PFNet18 with edge filters can outperform ResNet18 on some fine-grained image datasets. Furthermore, we show that the choice of filters matters, i.e. that edge filters lead to better recognition rates than random filters.

The concept of PFCNNs is fundamentally different from fine-tuning, where network weights are initialized with pre-trained weights. Our pre-defined Filter Convolutional Neural Networks (PFCNNs) apply pre-defined filters for all $n \times n$ convolution kernels with $n>1$, which are not changed during training. In our approach, we keep the pre-defined $1 \times n \times n$ convolution kernels and do not alter them during training. There are only a few different kernels (16 in our experiments) that are reused within each layer.

The paper is structured as follows:
Section~\ref{sec:related_work} gives a short overview of the related work. Section~\ref{sec:predefined_filter_networks} provides all details about the pre-defined filters and our architecture PFNet18. Section \ref{sec:experimental_setup} presents the training details and the image classification datasets.
The results are presented in Section~\ref{sec:results} and subsequently discussed in Section \ref{sec:discussion}.
The paper completes with our conclusions in Section~\ref{sec:conclusion}.


\section{Related work}
\label{sec:related_work}

The concept of pre-defined filters has a long history in vision. However, the limitation to fixed, pre-defined spatial filters seems to be novel in the intermediate layers of neural networks. In 1980 Fukushima \cite{fukushima_neocognitron_1980} used pre-defined filters in self-organizing neural networks. In today's era of deep networks, pre-defined filters are sometimes used as a pre-processing step to boost recognition performance. Ma et al. \cite{ma_pcfnet_2020} used pre-defined filters to incorporate domain knowledge into their training. They, however, replaced only the first convolutional layer of CNNs with trainable traditional image filters (Gabor filters).

Gavrikov and Keuper \cite{gavrikov_rethinking_2023} showed that learning linear combinations of fixed, random filters is a successful strategy to solve image classification problems, especially with wide CNNs (many channels). They also showed that the random filters can be shared across layers to further reduce the number of weights. In our study, we also learn linear combinations of filter outputs. However, we reduce the number of random filters to only 16 hand-picked kernels, that are applied in each depth-wise convolution operation. We show that the choice of these kernels matters, i.e. that edge filters outperform random kernels.


\section{Pre-defined Filter Networks}
\label{sec:predefined_filter_networks}

\begin{figure}[tb]
	\centering
	\includegraphics[trim=0.8cm 1cm 1cm 1cm, clip, width=0.8\linewidth]{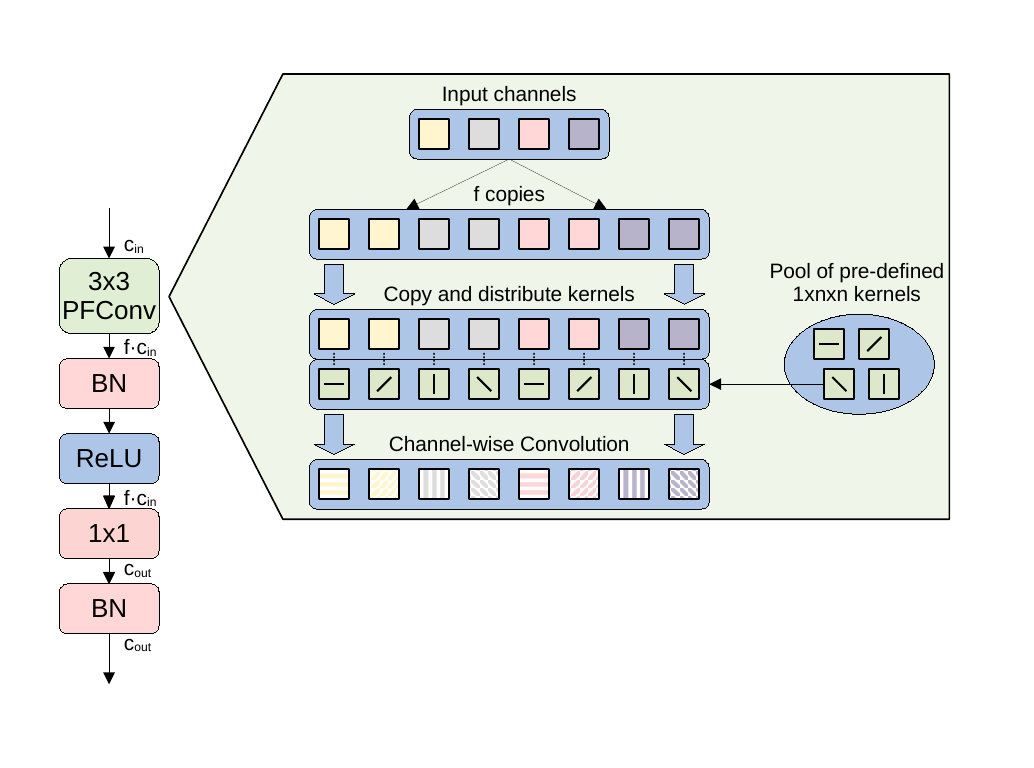}
	\caption{Pre-defined Filter Module (PFM). $1 \times n \times n$ kernels are taken from a small pool of kernels and are applied channel-wise as known from depthwise convolution. The order in which the kernels are distributed over the input channels is fixed.}
	\label{fig:PFM}
\end{figure}

\subsection{Pre-defined Filter Module}

The Pre-defined Filter Module (PFM) is illustrated in Figure \ref{fig:PFM}. It is structured like a depthwise convolutional layer. In the first part of the module, each input channel is convolved independently with exactly one $1 \times 3 \times 3$ filter kernel from a pool of $k$ pre-defined kernels. In our experiments, we use $k=16$ different edge kernels. The order in which the kernels are distributed over the input channels is fixed. The input channel with index $i$ is convolved with the kernel with index $i \mod k$.

The width of the intermediate part of the module can be chosen with the parameter f, as can be seen in Figure \ref{fig:PFM}. Conceptually, f determines the number of copies of the $n_\text{in}$ input channels prior to convolution. The number of intermediate channels after applying the pre-defined filter kernels is $n_\text{int} = n_\text{in} \cdot f$. Our implementation requires $n_\text{int} \mod{k} = 0$ with the number of pre-defined kernels $k$. Another requirement is $n_\text{int} \geq n_\text{in}$.

In the second part of the module, an $n_\text{int} \times 1 \times 1$ convolution creates an arbitrary number of linear combinations of the channels. By default, the first part of the PFM is frozen during training and only the weights in the $1 \times 1$ convolution are adjusted to the training set. 

\subsection{Choice of the parameter f}
The order in how the pre-defined filters are applied to the input channels should have no effect on the set of functions $F$ that can be learned by the network. The same should apply to a permutation of the channels in the input image, i.e., the order of kernels should not be part of the architecture. This has implications for the choice of the parameter $f$.

First, we discuss the first PFM that is applied to the RGB input image. For $f = k$, each kernel is applied once to each RGB input channel. In this case, the order of the input channels or the pre-defined filters is irrelevant. If we permute the input channels or the pre-defined filters we can always permute the weights in the consecutive $1 \times 1$ layer to get the same result as before.

Second, we discuss the second and following PFMs. Here, we choose $f = 1$. Again, the order of the pre-defined filters has no effect on the set of functions $F$ that can be learned by the network. If two input kernels at positions A and B are swapped, this operation can be undone in two steps. First, the preceding $1 \times 1$ layer swaps its output channels A and B by permuting its weights. If skip connections exist, all preceding PFMs have to adjust their $1 \times 1$ layer. Second, one needs to swap the inputs of the consecutive $1 \times 1$ layer, which can also be done by permuting the corresponding weights.

The order of kernels would matter for $f \in [2,k-1]$, which we avoid. This can easily be seen with some combinatorics. Please remember, that the input channel with index $i$ is convolved with the kernel $i \mod k$. As shown in Figure \ref{fig:PFM}, the input channels are copied prior to convolution. Thus, each input channel receives a set of $f$ different pre-defined kernels. However, if the order would be arbitrary we had $\binom{k}{f}$ filter combinations that could occur for a single input channel. This is larger than $k$ for $f \in [2,k-1]$. Therefore, if two pre-defined filters are swapped this could introduce new filter combinations, and the set of functions $F$ could change.

\begin{table}[tb]
\caption{Details of the PFNet18 architecture.}
\label{tab:PFNet18_architecture}
\centering
\begin{tabular}{ccc}
& & \\
\textbf{Layers} & \textbf{Output size} & \textbf{Kernel}\\
\midrule

\multirow{3}{*}{Pre-defined Filter Module} & 
\multirow{3}{*}{
    $\begin{bmatrix}
	3 \times 224 \times 224 \\
	48 \times 112 \times 112 \\
	64 \times 112 \times 112 \\
	\end{bmatrix}$ } &  
\multirow{3}{*}{
	$\begin{bmatrix}
	\text{input}, \\
	1 \times 3 \times 3, \text{depthwise} \\
	48 \times 1 \times 1, \text{pixelwise} \\
	\end{bmatrix}$ } \\
& & \\
& &	\\

\midrule

MaxPooling & $64 \times 56 \times 56$ &  $3 \times 3$, 64, stride 2 \\

\midrule

\multirow{2}{*}{Double HF Residual Module} & 
\multirow{2}{*}{
    $64 \times 56 \times 56$ } &  
\multirow{2}{*}{
	$\begin{bmatrix}
	1 \times 3 \times 3, \text{depthwise} \\
	64 \times 1 \times 1, \text{pixelwise} \\
	\end{bmatrix} \times 2$ } \\
& &	\\

\midrule

\multirow{2}{*}{Double HF Residual Module} & 
\multirow{2}{*}{
    $128 \times 28 \times 28$ } &  
\multirow{2}{*}{
	$\begin{bmatrix}
	1 \times 3 \times 3, \text{depthwise} \\
	64 \times 1 \times 1, \text{pixelwise} \\
	\end{bmatrix}$ } \\
& &	\\

\multirow{2}{*}{Double HF Residual Module} & 
\multirow{2}{*}{
    $128 \times 28 \times 28$ } &  
\multirow{2}{*}{
	$\begin{bmatrix}
	1 \times 3 \times 3, \text{depthwise} \\
	128 \times 1 \times 1, \text{pixelwise} \\
	\end{bmatrix}$ } \\
& &	\\

\midrule

\multirow{2}{*}{Double HF Residual Module} & 
\multirow{2}{*}{
    $256 \times 14 \times 14$ } &  
\multirow{2}{*}{
	$\begin{bmatrix}
	1 \times 3 \times 3, \text{depthwise} \\
	128 \times 1 \times 1, \text{pixelwise} \\
	\end{bmatrix}$ } \\
& &	\\

\multirow{2}{*}{Double HF Residual Module} & 
\multirow{2}{*}{
    $256 \times 14 \times 14$ } &  
\multirow{2}{*}{
	$\begin{bmatrix}
	1 \times 3 \times 3, \text{depthwise} \\
	256 \times 1 \times 1, \text{pixelwise} \\
	\end{bmatrix}$ } \\
& &	\\

\midrule
\multirow{2}{*}{Double HF Residual Module} & 
\multirow{2}{*}{
    $512 \times 7 \times 7$ } &  
\multirow{2}{*}{
	$\begin{bmatrix}
	1 \times 3 \times 3, \text{depthwise} \\
	256 \times 1 \times 1, \text{pixelwise} \\
	\end{bmatrix}$ } \\
& &	\\

\multirow{2}{*}{Double HF Residual Module} & 
\multirow{2}{*}{
    $512 \times 7 \times 7$ } &  
\multirow{2}{*}{
	$\begin{bmatrix}
	1 \times 3 \times 3, \text{depthwise} \\
	512 \times 1 \times 1, \text{pixelwise} \\
	\end{bmatrix}$ } \\
& &	\\

\midrule		
\multirow{2}{*}{Classification layer} & $512 \times 1 \times 1$ &  Adaptive average pool \\
\cmidrule(l){2-3}
& & fully connected, softmax \\
\end{tabular}
\end{table}

\begin{table}[tb]
\caption{Details of the ResNet18 architecture.}
\label{tab:ResNet18_architecture}
\centering
\begin{tabular}{ccc}
& & \\
\textbf{Layers} & \textbf{Output size} & \textbf{Kernel}\\
\midrule

Convolution & $64 \times 112 \times 112$ &  $7 \times 7$, 64, stride 2 \\
\midrule
MaxPooling & $64 \times 56 \times 56$ &  $3 \times 3$, 64, stride 2 \\

\midrule
\multirow{2}{*}{Basic Block} & \multirow{2}{*}{$64 \times 56 \times 56$} &  
\multirow{2}{*}{
	$\begin{bmatrix}
	64 \times 3 \times 3 \\
	64 \times 3 \times 3
	\end{bmatrix} \times 2$ } \\
& & \\

\midrule

\multirow{2}{*}{Basic Block} & \multirow{2}{*}{$128 \times 28 \times 28$} &  
\multirow{2}{*}{
	$\begin{bmatrix}
	64 \times 3 \times 3\\
	128 \times 3 \times 3
	\end{bmatrix}$ } \\
& & \\

\multirow{2}{*}{Basic Block} & \multirow{2}{*}{$128 \times 28 \times 28$} &  
\multirow{2}{*}{
	$\begin{bmatrix}
	128 \times 3 \times 3\\
	128 \times 3 \times 3
	\end{bmatrix}$ } \\
& & \\

\midrule

\multirow{2}{*}{Basic Block} & \multirow{2}{*}{$256 \times 14 \times 14$} &  
\multirow{2}{*}{
	$\begin{bmatrix}
	128 \times 3 \times 3\\
	256 \times 3 \times 3
	\end{bmatrix}$ } \\
& & \\

\multirow{2}{*}{Basic Block} & \multirow{2}{*}{$256 \times 14 \times 14$} &  
\multirow{2}{*}{
	$\begin{bmatrix}
	256 \times 3 \times 3\\
	256 \times 3 \times 3
	\end{bmatrix}$ } \\
& & \\

\midrule

\multirow{2}{*}{Basic Block} & \multirow{2}{*}{$512 \times 7 \times 7$} &  
\multirow{2}{*}{
	$\begin{bmatrix}
	256 \times 3 \times 3\\
	512 \times 3 \times 3
	\end{bmatrix}$ } \\
& & \\

\multirow{2}{*}{Basic Block} & \multirow{2}{*}{$512 \times 7 \times 7$} &  
\multirow{2}{*}{
	$\begin{bmatrix}
	512 \times 3 \times 3\\
	512 \times 3 \times 3
	\end{bmatrix}$ } \\
& & \\

\midrule		
\multirow{2}{*}{Classification layer} & $512 \times 1 \times 1$ &  Adaptive average pool \\
\cmidrule(l){2-3}
& & fully connected, softmax \\
\end{tabular}
\end{table}

\subsection{PFNet18 architecture}

The PFNet18 architecture is described in Table \ref{tab:PFNet18_architecture}. For comparison, ResNet18 architecture is presented in Table~\ref{tab:ResNet18_architecture}. Starting from ResNet18, we replace the convolution layers with Pre-defined Filter Modules. 
The first module has $f = 16$, which is also the number of pre-defined kernels. Thus, the first layer has $n_\text{int} = 3 \cdot 16 = 48$ intermediate channels. To be consistent with ResNet18, we choose $n_\text{out} = 64$ output channels for the $1 \times 1$ convolution. 
After the first PFM, we stack 8 residually connected Basic Blocks and replace the convolution operations with PFMs with $f = 1$. The residual connections are kept. After the last block, there is an adaptive average pooling layer and a fully connected layer just like in ResNet18.
PFNet18 has 1.46 million parameters, which is only $13\%$ of the 11.23 million parameters of the ResNet18.

\subsection{Choice of pre-defined filters}

\begin{figure}[tb]
	\centering
	\includegraphics[trim=1cm 1cm 1cm 1cm, clip, width=0.6\linewidth]{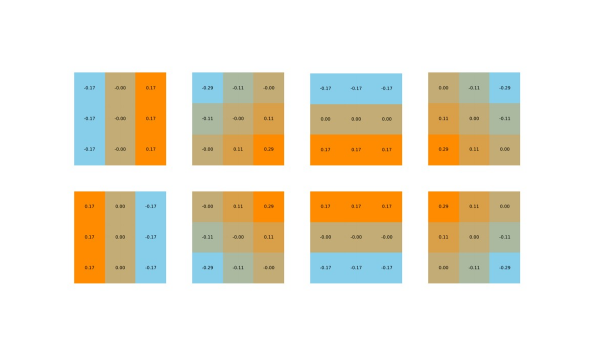} 
	\\[0.5cm]
	\includegraphics[trim=1cm 1cm 1cm 1cm, clip, width=0.6\linewidth]{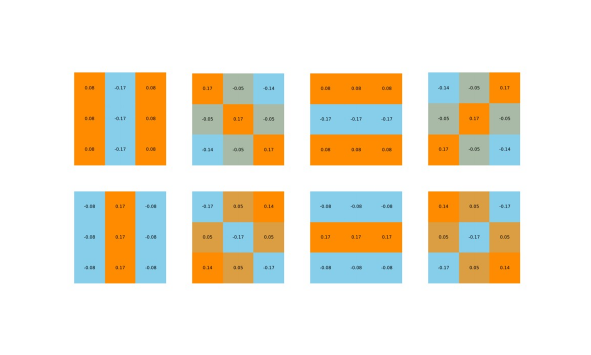}
	\caption{8 uneven and 8 even $1 \times 3 \times 3$ convolution kernels used in PFNet18.}
    \label{fig:kernels}
\end{figure}

In our experiments, we employ only 16 different pre-defined $1 \times 3 \times 3$ filters. We choose edge filters because they provide gradient information and are often found in rudimentary forms in trained CNNs \cite{krizhevsky_imagenet_2012, gavrikov_cnn_2022}. Figure~\ref{fig:kernels} presents 8 uneven and 8 even edge filters in different orientations including horizontal, vertical, and diagonal. We suppose that employing edge filters in PFCNNs will lead to an appropriate bias that will help the network to learn robust features. The kernel elements $w_i$ are normalized such that $\sum_{i} w_i = 0$ and $\sum_{i} |w_i| = 1$. The constant component is missing in the convolution kernels, which should bias the network toward the processing of edges and shapes. As there are only 8 linearly independent kernels (some of the kernels have a flipped sign) the kernels span an 8-dimensional space.

\section{Experimental setup}
\label{sec:experimental_setup}

We train and test PFNet18 on several image classification datasets and compare its performance with ResNet18. The following sections present the benchmark datasets and all training details.

\subsection{Datasets}

\begin{table}[tb]
\caption{Images from the benchmark datasets showing each two instances of four classes.}
\label{tab:dataset_examples}
\centering
\begin{adjustbox}{width=\linewidth}
\begin{tabular}{cc}

& \\
Caltech101 & \begin{minipage}[c]{13cm}
\includegraphics[width=1.5cm, height=1.5cm]{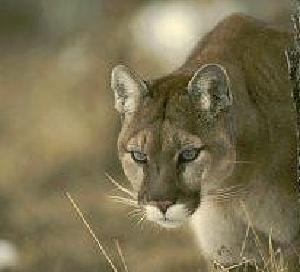}
\includegraphics[width=1.5cm, height=1.5cm]{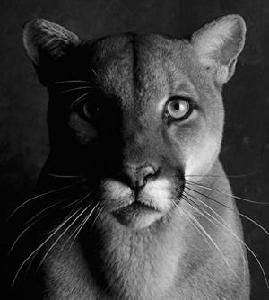}
\includegraphics[width=1.5cm, height=1.5cm]{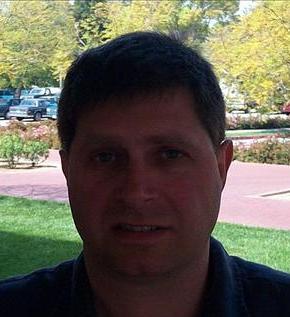}
\includegraphics[width=1.5cm, height=1.5cm]{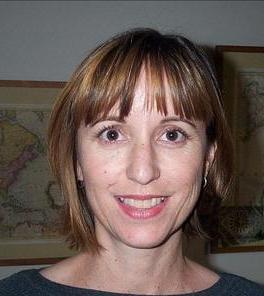}
\includegraphics[width=1.5cm, height=1.5cm]{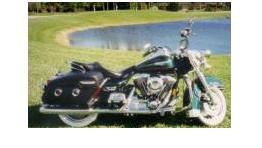}
\includegraphics[width=1.5cm, height=1.5cm]{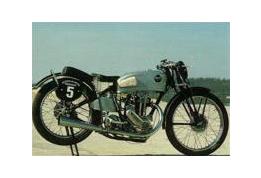}
\includegraphics[width=1.5cm, height=1.5cm]{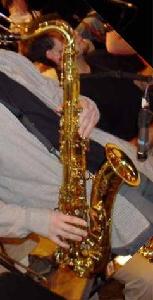}
\includegraphics[width=1.5cm, height=1.5cm]{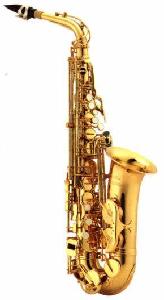}
\end{minipage}
\\

CIFAR10 & \begin{minipage}[c]{13cm}
\includegraphics[width=1.5cm, height=1.5cm]{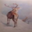}
\includegraphics[width=1.5cm, height=1.5cm]{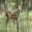}
\includegraphics[width=1.5cm, height=1.5cm]{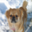}
\includegraphics[width=1.5cm, height=1.5cm]{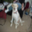}
\includegraphics[width=1.5cm, height=1.5cm]{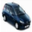}
\includegraphics[width=1.5cm, height=1.5cm]{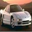}
\includegraphics[width=1.5cm, height=1.5cm]{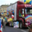}
\includegraphics[width=1.5cm, height=1.5cm]{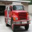}
\end{minipage}
\\

CUB-200-2011 & \begin{minipage}[c]{13cm}
\includegraphics[width=1.5cm, height=1.5cm]{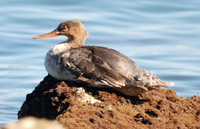}
\includegraphics[width=1.5cm, height=1.5cm]{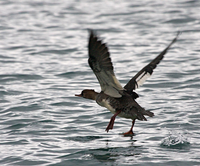}
\includegraphics[width=1.5cm, height=1.5cm]{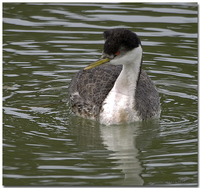}
\includegraphics[width=1.5cm, height=1.5cm]{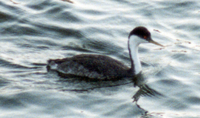}
\includegraphics[width=1.5cm, height=1.5cm]{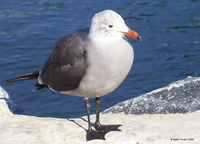}
\includegraphics[width=1.5cm, height=1.5cm]{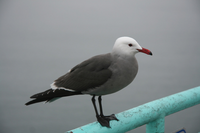}
\includegraphics[width=1.5cm, height=1.5cm]{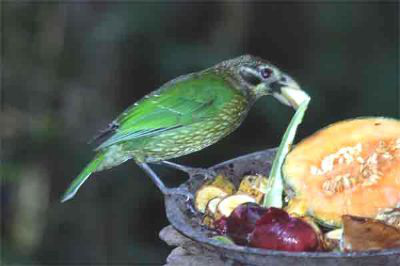}
\includegraphics[width=1.5cm, height=1.5cm]{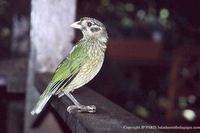}
\end{minipage}
\\

FGVC-Aircraft & \begin{minipage}[c]{13cm}
\includegraphics[width=1.5cm, height=1.5cm]{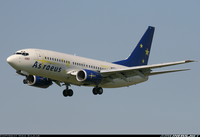}
\includegraphics[width=1.5cm, height=1.5cm]{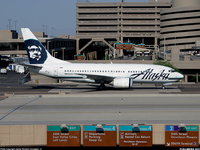}
\includegraphics[width=1.5cm, height=1.5cm]{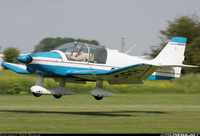}
\includegraphics[width=1.5cm, height=1.5cm]{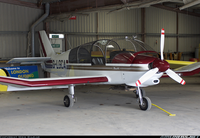}
\includegraphics[width=1.5cm, height=1.5cm]{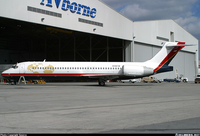}
\includegraphics[width=1.5cm, 
height=1.5cm]{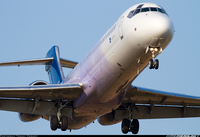}
\includegraphics[width=1.5cm, height=1.5cm]{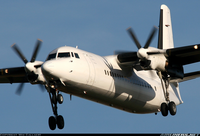}
\includegraphics[width=1.5cm, height=1.5cm]{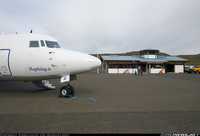}
\end{minipage}
\\

Flowers102 & \begin{minipage}[c]{13cm}
\includegraphics[width=1.5cm, height=1.5cm]{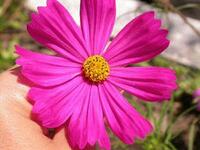}
\includegraphics[width=1.5cm, height=1.5cm]{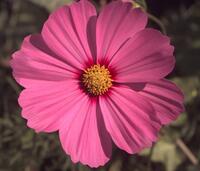}
\includegraphics[width=1.5cm, height=1.5cm]{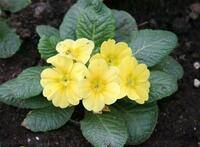}
\includegraphics[width=1.5cm, height=1.5cm]{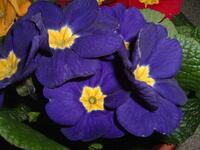}
\includegraphics[width=1.5cm, height=1.5cm]{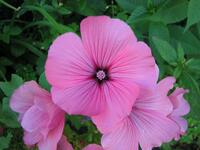}
\includegraphics[width=1.5cm, height=1.5cm]{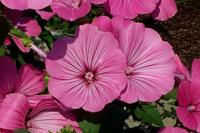}
\includegraphics[width=1.5cm, height=1.5cm]{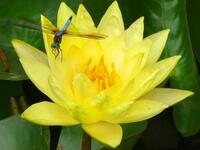}
\includegraphics[width=1.5cm, height=1.5cm]{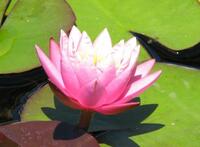}
\end{minipage}
\\

Stanford Cars & \begin{minipage}[c]{13cm}
\includegraphics[width=1.5cm, height=1.5cm]{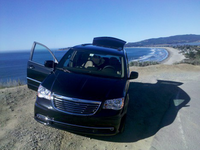}
\includegraphics[width=1.5cm, height=1.5cm]{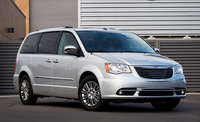}
\includegraphics[width=1.5cm, height=1.5cm]{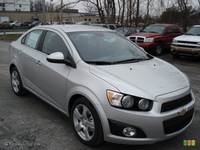}
\includegraphics[width=1.5cm, height=1.5cm]{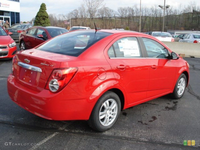}
\includegraphics[width=1.5cm, height=1.5cm]{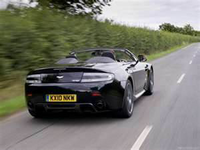}
\includegraphics[width=1.5cm, height=1.5cm]{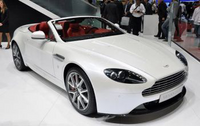}
\includegraphics[width=1.5cm, height=1.5cm]{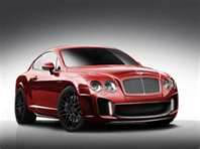}
\includegraphics[width=1.5cm, height=1.5cm]{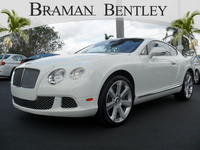}
\end{minipage}
\\

\end{tabular}
\end{adjustbox}
\end{table}

Example images for each dataset are presented in Table \ref{tab:dataset_examples}. 
The Caltech101 dataset \cite{li_fei_fei_learning_2004} has a total of 8677 images, which were collected from the internet. They are scaled to be about 300 pixels wide. The dataset contains 101 categories, as well as a background class, which is ignored in the experiments. The categories are quite diverse and reach from animals and plants to electronic products and vehicles. The objects are shown in cluttered environments or in front of realistic or white backgrounds. As no official split is available, we randomly pick 20 training images and 10 test images per class.

The Caltech-UCSD Birds-200-2011 dataset (CUB-200-2011) \cite{wah_caltech_ucsd_2011} originated from the Caltech-UCSD Birds 200 dataset \cite{welinder_caltech_ucsd_2010} from 2010.
We use the official split with 5994 training and 5794 test images. The average width of the images is about 470 pixels. There are 200 challenging bird categories (classes), which have plenty of inter-class variation due to deformations, plumage color, lighting, perspective, and pose.

The Fine-Grained Visual Classification of Aircraft dataset (FGVC-Aircraft) \cite{maji_fine_grained_2013} provides 6667 images for training and 3333 images for testing. It contains large images with an average width of 1100 pixels. We use the 100 categories of airplane models. The airplanes are rigid and do not deform. However, the same airplane model can appear quite different depending on the advertisement, airlines, perspective, and cropping.

The images from the 102 Category Flower dataset (Flowers102) \cite{nilsback_automated_2008} show one or several blossoms of 102 flower types. As the other datasets in our study have training and test sets only, we merge the official training and validation set of Flowers102 to our training set and we test on the official test set. Accordingly, the training set contains 2040 images, and the test set 6149 images. The images have a mean width of 630 pixels. While some flower types can have subtle differences in their appearance, there are large intra-class variations due to scaling, perspective, lighting, and color variants. We only use categorical information within the dataset.

The StanfordCars dataset \cite{krause_3d_2013} from Stanford University provides 8144 training and 8041 test images. The average width of an image is 700 pixels. The dataset covers 196 different car models. Usually, one car is shown per image. The cars can be on the road or indoors, shown from the front, side, or back, introducing many variations to the dataset.

CIFAR10 \cite{krizhevsky_learning_2009} contains 50000 training images and 10000 test images of the size $32 \times 32$ pixels. The dataset provides the 10 classes plane, car, bird, cat, deer, dog, frog, horse, ship, and truck. 
For our experiments on the CIFAR10 dataset, we adjusted the architectures to be compatible with small image shapes by removing the max pool layer and having a stride of 1 in the first convolutional layer and a kernel size of 3. These adjustments apply for both, PFNet18 and ResNet18.

\subsection{Training details}

The network weights are initialized using Kaiming normal initialization \cite{he_delving_2015}. We apply the Lamb optimizer \cite{you_large_2020} to minimize the cross-entropy loss of our models. The batch size is 64. We train for 300 epochs. The initial learning rate of $\text{lr}_\text{init} = 0.003$ is step-wise reduced to $\text{lr}_\text{init} \cdot 10^{-2}$. The weight decay is 1.
We use the Pytorch framework  \cite{paszke_pytorch_2019} to implement the models and to perform the training on an NVidia GTX 1080 Ti. 
The experiments are repeated 5 times with random seeds. Each seed affects the weight initialization of the networks, the mini-batch aggregation, and random effects during data augmentation.
For image augmentation, we apply random cropping and random horizontal flipping.


\section{Results}
\label{sec:results}

\subsection{Benchmarks}

\begin{table}[tb]
\caption{Test accuracy on benchmark datasets for training from scratch and 5 different seeds.}
\label{tab:datasets_results}
\centering

\begin{tabular}{ccc}
& & \\
Dataset & PFNet18 & ResNet18  \\
\midrule
Caltech101 & \textbf{65.60$\pm$0.66} & 57.19$\pm$0.78  \\
CIFAR10 & 92.15$\pm$0.18 & \textbf{94.33$\pm$0.11} \\
CUB-200-2011 & 53.00$\pm$0.55 & \textbf{58.51$\pm$0.53}  \\
FGVC-Aircraft &\textbf{75.49$\pm$0.29} & 73.32$\pm$1.06 \\
Flowers102 & \textbf{80.66$\pm$0.35} & 73.40$\pm$0.34 \\
Stanford Cars & 77.06$\pm$0.42 & \textbf{77.90$\pm$0.37} \\
\end{tabular}
\end{table}

The average test performance of the PFNet18 and ResNet18 models is presented in Table \ref{tab:datasets_results}. PFNet18 outperforms ResNet18 on the Caltech101, FGVC-Aircraft, and Flowers102 datasets. On Caltech101, PFNet18 has an almost $10\%$ higher test accuracy than ResNet18 and on Flowers102 the improvement is $7\%$. On the Stanford Cars dataset, both architectures perform very similarly. ResNet18 achieves higher accuracies on the CIFAR10 and the CUB-200-2011 datasets. 
The experiments show that the restriction to employ only 16 different pre-defined $1 \times 3 \times 3$ filters is sufficient to learn complex relationships in image data. The PFNet18 models reached high accuracy by simply finding linear combinations of pre-defined filter outputs. This is intriguing because it demonstrates that the spatial kernels do not have to be learned at all in many cases.

\subsection{Feature visualization}

\begin{table*}[p]
\caption{Feature visualization for models trained on the Caltech101 and Flowers102 dataset. Each image is an input that maximizes the activation of a specific channel in a specific layer. The visualized classes are cougar face (top left), face easy (top right), motorbikes (bottom left), and saxophone (bottom right) for the Caltech101 dataset and Mexican aster (top left), primula (top right), tree mallow (bottom left) and water lily (bottom right) for the Flowers102 dataset. Best viewed with zoom.}
\label{tab:FeatureVisualization}
\centering
\begin{adjustbox}{width=0.8\linewidth}
\begin{tabular}{ccccc}

\textbf{Layer}	& 
\textbf{Caltech101, PFNet18} & 
\textbf{Caltech101, ResNet18} & 
\textbf{Flowers102, PFNet18} & 
\textbf{Flowers102, ResNet18} \\

\midrule

Block 1 &
\begin{minipage}[c]{\featvisminisize}
\includegraphics[width=\featvisimgsize]{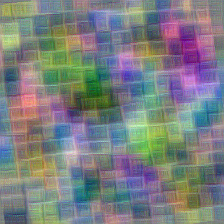}
\includegraphics[width=\featvisimgsize]{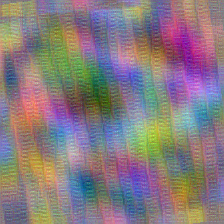}
\\
\includegraphics[width=\featvisimgsize]{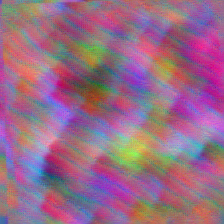}
\includegraphics[width=\featvisimgsize]{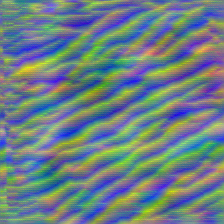}
\\
\end{minipage}
&
\begin{minipage}[c]{\featvisminisize}
\includegraphics[width=\featvisimgsize]{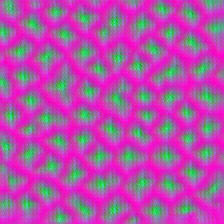}
\includegraphics[width=\featvisimgsize]{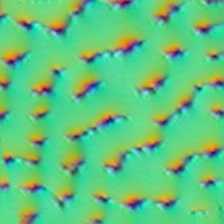}
\\
\includegraphics[width=\featvisimgsize]{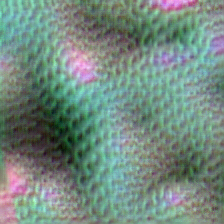}
\includegraphics[width=\featvisimgsize]{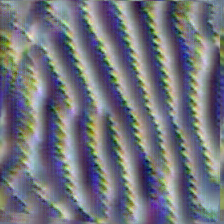}
\\
\end{minipage}
&
\begin{minipage}[c]{\featvisminisize}
\includegraphics[width=\featvisimgsize]{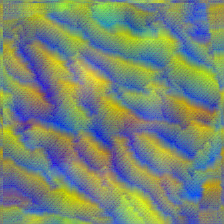}
\includegraphics[width=\featvisimgsize]{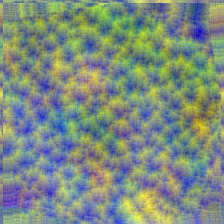}
\\
\includegraphics[width=\featvisimgsize]{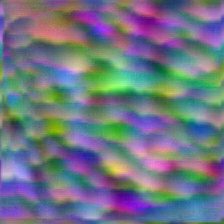}
\includegraphics[width=\featvisimgsize]{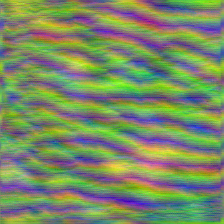}
\\
\end{minipage}
&
\begin{minipage}[c]{\featvisminisize}
\includegraphics[width=\featvisimgsize]{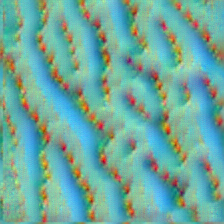}
\includegraphics[width=\featvisimgsize]{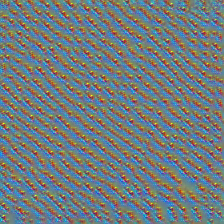}
\\
\includegraphics[width=\featvisimgsize]{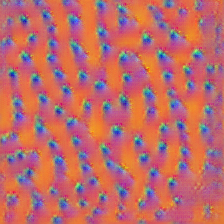}
\includegraphics[width=\featvisimgsize]{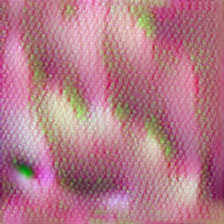}
\\
\end{minipage}
\\

\midrule

Block 2 &
\begin{minipage}[c]{\featvisminisize}
\includegraphics[width=\featvisimgsize]{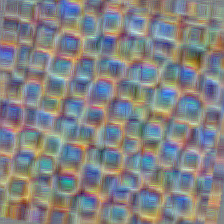}
\includegraphics[width=\featvisimgsize]{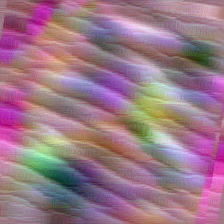}
\\
\includegraphics[width=\featvisimgsize]{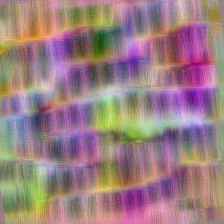}
\includegraphics[width=\featvisimgsize]{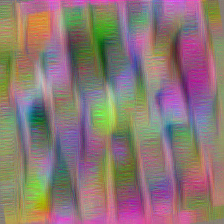}
\\
\end{minipage}
&
\begin{minipage}[c]{\featvisminisize}
\includegraphics[width=\featvisimgsize]{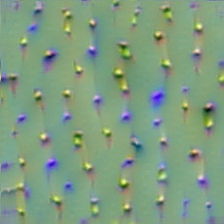}
\includegraphics[width=\featvisimgsize]{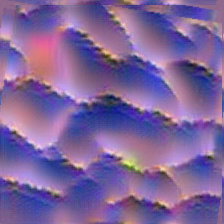}
\\
\includegraphics[width=\featvisimgsize]{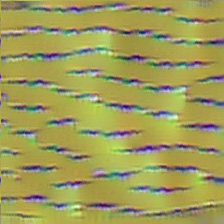}
\includegraphics[width=\featvisimgsize]{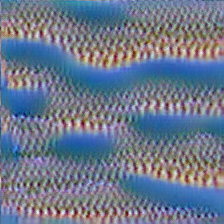}
\\
\end{minipage}
&
\begin{minipage}[c]{\featvisminisize}
\includegraphics[width=\featvisimgsize]{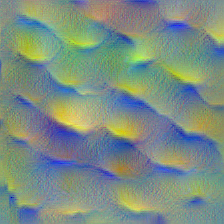}
\includegraphics[width=\featvisimgsize]{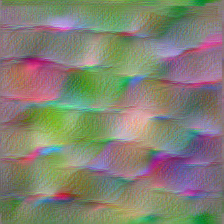}
\\
\includegraphics[width=\featvisimgsize]{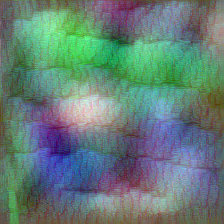}
\includegraphics[width=\featvisimgsize]{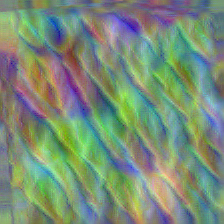}
\\
\end{minipage}
&
\begin{minipage}[c]{\featvisminisize}
\includegraphics[width=\featvisimgsize]{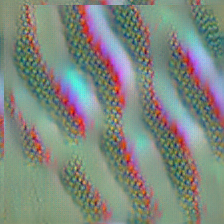}
\includegraphics[width=\featvisimgsize]{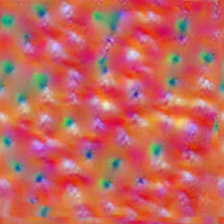}
\\
\includegraphics[width=\featvisimgsize]{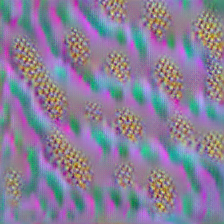}
\includegraphics[width=\featvisimgsize]{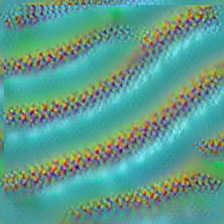}
\\
\end{minipage}
\\

\midrule

Block 3 &
\begin{minipage}[c]{\featvisminisize}
\includegraphics[width=\featvisimgsize]{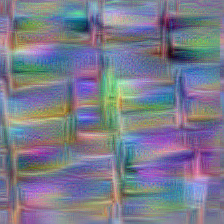}
\includegraphics[width=\featvisimgsize]{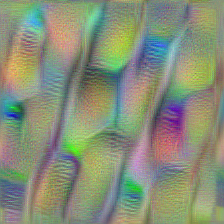}
\\
\includegraphics[width=\featvisimgsize]{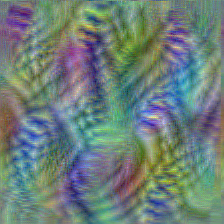}
\includegraphics[width=\featvisimgsize]{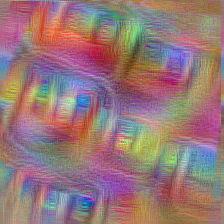}
\\
\end{minipage}
&
\begin{minipage}[c]{\featvisminisize}
\includegraphics[width=\featvisimgsize]{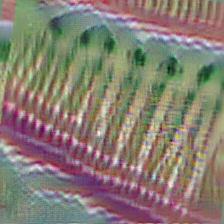}
\includegraphics[width=\featvisimgsize]{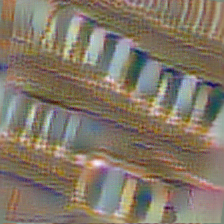}
\\
\includegraphics[width=\featvisimgsize]{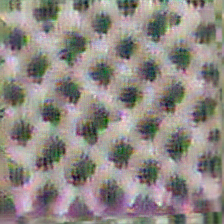}
\includegraphics[width=\featvisimgsize]{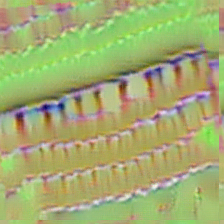}
\\
\end{minipage}
&
\begin{minipage}[c]{\featvisminisize}
\includegraphics[width=\featvisimgsize]{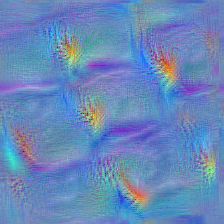}
\includegraphics[width=\featvisimgsize]{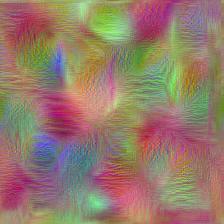}
\\
\includegraphics[width=\featvisimgsize]{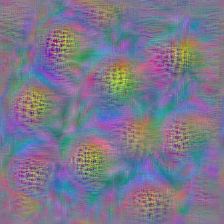}
\includegraphics[width=\featvisimgsize]{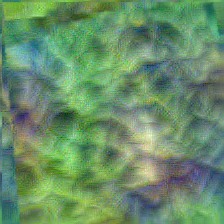}
\\
\end{minipage}
&
\begin{minipage}[c]{\featvisminisize}
\includegraphics[width=\featvisimgsize]{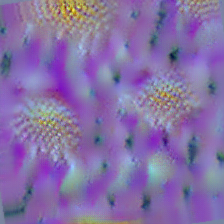}
\includegraphics[width=\featvisimgsize]{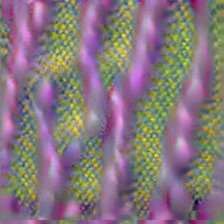}
\\
\includegraphics[width=\featvisimgsize]{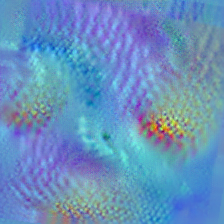}
\includegraphics[width=\featvisimgsize]{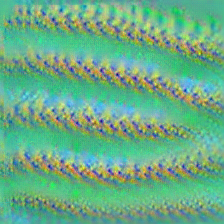}
\\
\end{minipage}
\\

\midrule

Block 4 &
\begin{minipage}[c]{\featvisminisize}
\includegraphics[width=\featvisimgsize]{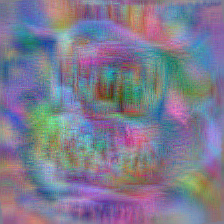}
\includegraphics[width=\featvisimgsize]{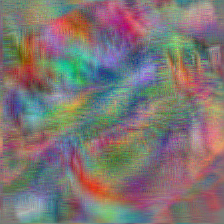}
\\
\includegraphics[width=\featvisimgsize]{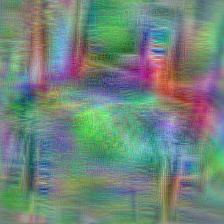}
\includegraphics[width=\featvisimgsize]{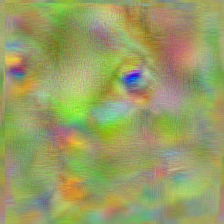}
\\
\end{minipage}
&
\begin{minipage}[c]{\featvisminisize}
\includegraphics[width=\featvisimgsize]{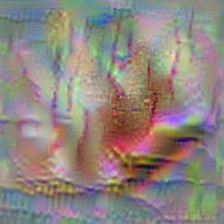}
\includegraphics[width=\featvisimgsize]{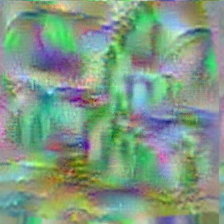}
\\
\includegraphics[width=\featvisimgsize]{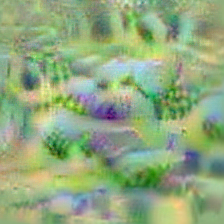}
\includegraphics[width=\featvisimgsize]{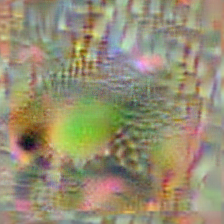}
\\
\end{minipage}
&
\begin{minipage}[c]{\featvisminisize}
\includegraphics[width=\featvisimgsize]{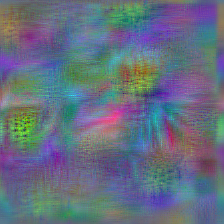}
\includegraphics[width=\featvisimgsize]{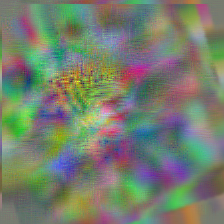}
\\
\includegraphics[width=\featvisimgsize]{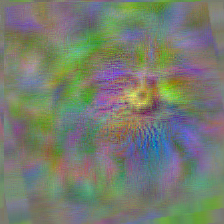}
\includegraphics[width=\featvisimgsize]{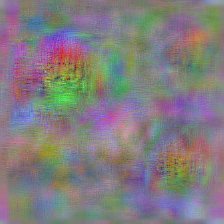}
\\
\end{minipage}
&
\begin{minipage}[c]{\featvisminisize}
\includegraphics[width=\featvisimgsize]{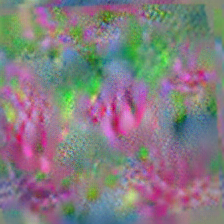}
\includegraphics[width=\featvisimgsize]{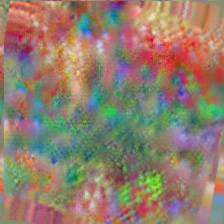}
\\
\includegraphics[width=\featvisimgsize]{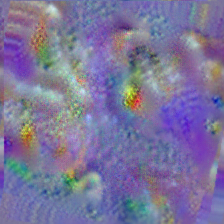}
\includegraphics[width=\featvisimgsize]{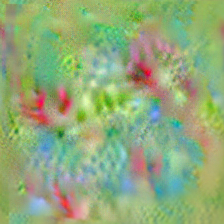}
\\
\end{minipage}
\\

\midrule

Classes &
\begin{minipage}[c]{\featvisminisize}
\includegraphics[width=\featvisimgsize]{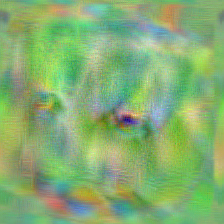}
\includegraphics[width=\featvisimgsize]{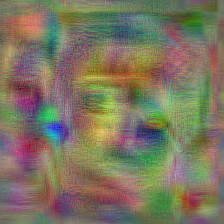}
\\
\includegraphics[width=\featvisimgsize]{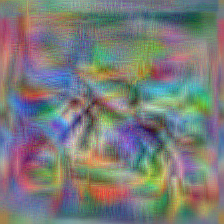}
\includegraphics[width=\featvisimgsize]{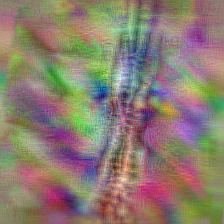}
\\
\end{minipage}
&
\begin{minipage}[c]{\featvisminisize}
\includegraphics[width=\featvisimgsize]{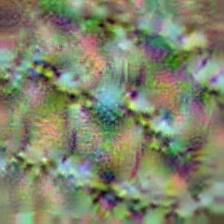}
\includegraphics[width=\featvisimgsize]{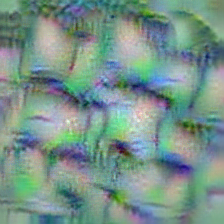}
\\
\includegraphics[width=\featvisimgsize]{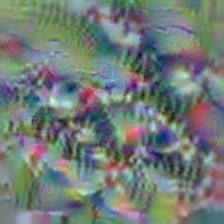}
\includegraphics[width=\featvisimgsize]{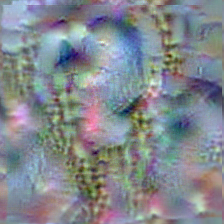}
\\
\end{minipage}
&
\begin{minipage}[c]{\featvisminisize}
\includegraphics[width=\featvisimgsize]{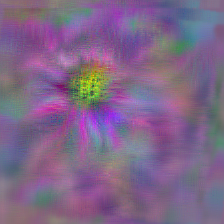}
\includegraphics[width=\featvisimgsize]{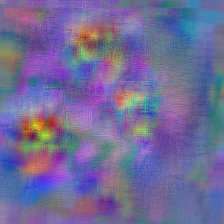}
\\
\includegraphics[width=\featvisimgsize]{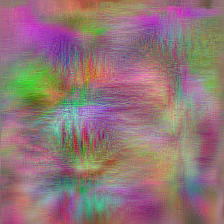}
\includegraphics[width=\featvisimgsize]{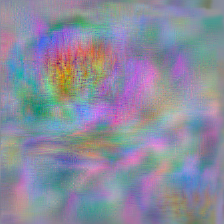}
\\
\end{minipage}
&
\begin{minipage}[c]{\featvisminisize}
\includegraphics[width=\featvisimgsize]{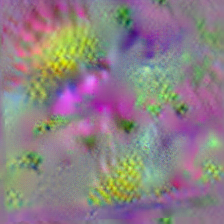}
\includegraphics[width=\featvisimgsize]{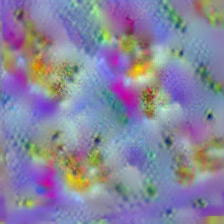}
\\
\includegraphics[width=\featvisimgsize]{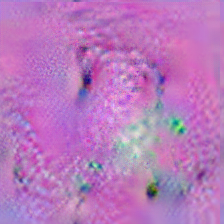}
\includegraphics[width=\featvisimgsize]{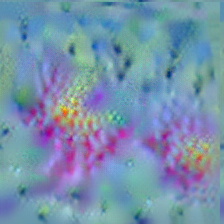}
\\
\end{minipage}
\\

\midrule

Class reference &
\begin{minipage}[c]{\featvisminisize}
\includegraphics[width=\featvisimgsize, height=\featvisimgsize]{FIG_Caltech101_cougar_face_image_0011.jpg}
\includegraphics[width=\featvisimgsize, height=\featvisimgsize]{FIG_Caltech101_Faces_easy_image_0015.jpg}
\\
\includegraphics[width=\featvisimgsize, height=\featvisimgsize]{FIG_Caltech101_Motorbikes_image_0283.jpg}
\includegraphics[width=\featvisimgsize, height=\featvisimgsize]{FIG_Caltech101_saxophone_image_0014.jpg}
\\
\end{minipage}
&
\begin{minipage}[c]{\featvisminisize}
\includegraphics[width=\featvisimgsize, height=\featvisimgsize]{FIG_Caltech101_cougar_face_image_0049.jpg}
\includegraphics[width=\featvisimgsize, height=\featvisimgsize]{FIG_Caltech101_Faces_easy_image_0083.jpg}
\\
\includegraphics[width=\featvisimgsize, height=\featvisimgsize]{FIG_Caltech101_Motorbikes_image_0705.jpg}
\includegraphics[width=\featvisimgsize, height=\featvisimgsize]{FIG_Caltech101_saxophone_image_0028.jpg}
\\
\end{minipage}
&
\begin{minipage}[c]{\featvisminisize}
\includegraphics[width=\featvisimgsize, height=\featvisimgsize]{FIG_Flowers_mexican_aster_image_06942.jpg}
\includegraphics[width=\featvisimgsize, height=\featvisimgsize]{FIG_Flowers_primula_image_03731.jpg}
\\
\includegraphics[width=\featvisimgsize, height=\featvisimgsize]{FIG_Flowers_tree_mallow_image_02917.jpg}
\includegraphics[width=\featvisimgsize, height=\featvisimgsize]{FIG_Flowers_water_lily_image_00274.jpg}
\\
\end{minipage}
&
\begin{minipage}[c]{\featvisminisize}
\includegraphics[width=\featvisimgsize, height=\featvisimgsize]{FIG_Flowers_mexican_aster_image_06948.jpg}
\includegraphics[width=\featvisimgsize, height=\featvisimgsize]{FIG_Flowers_primula_image_03729.jpg}
\\
\includegraphics[width=\featvisimgsize, height=\featvisimgsize]{FIG_Flowers_tree_mallow_image_02918.jpg}
\includegraphics[width=\featvisimgsize, height=\featvisimgsize]{FIG_Flowers_water_lily_image_00271.jpg}
\\
\end{minipage}
\\

\end{tabular}
\end{adjustbox}
\end{table*}

The promising results indicate that our choice of pre-defined filters introduces a good bias for at least half of the datasets. In an ablation study in Section \ref{sec:results:ablation_study} we study the choice of filters quantitatively. In this section, we study the learned filters of PFNet18 qualitatively using the feature visualization technique \cite{erhan_visualizing_2009}. 
Feature visualization reveals specific input patterns that maximize the activation of specific network units. The outputs of this technique are input images that show characteristics of the features processed within the network. The idea is to initialize an input image with Gaussian noise and to modify the image such that it maximizes the activation of a specific neuron using gradient ascent. Hence, this procedure is also called activation maximization. 
Let $\theta$ be the network weights and let $f_{ij}(\theta, x)$ be the jth feature map in layer i given $\theta$ and some input image $x$: for activation maximization, the input image $x^*$ with
\begin{equation*}
    x^* = \argmax_{x \in \Re^{C \times H \times W}} f_{ij}(\theta, x)
\end{equation*}
is determined \cite{erhan_visualizing_2009}. We perform the optimization using gradient ascent on a fixed $\theta$. Such generated images often look unnatural and regularization approaches have been used in the literature to improve the visual quality \cite{mahendran_understanding_2015}. Between the update steps $x \longleftarrow x + \mu \nabla_x f_{ij}(\theta, x)$ with learning rate $\mu$ we apply random image transformations to support the gradient ascent to find robust maxima. Our transformations include random rotation, scaling, blurring, cropping, pixel rolling, and shifting/scaling the image tensor distribution toward the normal distribution. Similar transformations have been applied in the works \cite{yosinski_understanding_2015, mordvintsev_inceptionism_2015, nguyen_deep_2015}. 

Table \ref{tab:FeatureVisualization} shows feature visualizations of PFNet18 and ResNet18 trained on the Caltech101 and Flowers102 datasets. Table \ref{tab:FeatureVisualization} is best viewed digitally with zoom. The images were picked by hand to illustrate the variation and the characteristics of the features in different layers.

The results indicate that both PFNet18 and ResNet18 are able to learn complex visual features. One observes that the features processed at the end of the first block consist of simple, repetitive textures. At the end of the second block, more details are added to these textures and it is possible to see a difference between the datasets. The models that were trained on the Flowers102 dataset produce leaf and flower prototypes. The models that were trained on Caltech101 also include rectangular shapes and more variety in general. Similar observations apply for the third and fourth blocks, where the first object instances appear. The feature visualization of the classification layer reveals what the networks expect to look like cougar face, face, motorbike, and saxophone (Caltech101) and Mexican aster, primula, tree mallow, and water lily (Flowers102).

Feature visualization reveals that both PFNet18 and ResNet18 learned object-specific features. On the Caltech101 and Flowers102 datasets, the PFNets18's feature visualizations look more convincing. On Caltech101, exactly one instance of the target class is generated with a plausible shape. The cougar's face consists of a head with two eyes and ears. The human face has a nose, two eyes (the second one is difficult to see), hair, and a flat chin. The motorbike has two wheels, a saddle, and a handle. The saxophone is a long stick with buttons and a shimmering surface. ResNet18 does not generate such clear shapes and seems to focus more on textures. Among many feature visualizations and also the other classes, no examples could be found that look as convincing as those of PFNet18. The results show that PFCNNs are able to learn complex, object-specific features from only 20 training images per class.

\subsection{Computational efficiency}

\begin{table}[tb]
\caption{Computational efficiency of PFNet18 and ResNet18 considering model size and speed. FP denotes forward pass and BP denotes backward pass. During BP the hyperparameters described above are used. The input tensors have the shape $64 \times 3 \times 224 \times 224$.}
\label{tab:model_size}
\centering
\begin{adjustbox}{width=\linewidth}
\begin{tabular}{ccccccc}
Model     & Parameters ($10^6$) & Size (MB) & FP (ms) & BP (ms) & GPU Memory FP (GB) & Mult-Adds ($10^9$) \\
\midrule
PFNet18     & 1.46 & 6  & 37 & 70 & 3.9 & 0.26 \\
ResNet18   & 11.23 & 45 & 27 & 65 & 3.0 & 1.81 \\
\end{tabular}
\end{adjustbox}
\end{table}

Computational efficiency and model size are important factors for training and deploying deep networks, especially on limited hardware with harsh energy and memory requirements. Table \ref{tab:model_size} summarizes relevant computational aspects of the models.
PFNet18 has only 1.46 million parameters and requires only 6 MB space on the disk, whereas ResNet18 needs 45 MB and has 11.23 million parameters. The reduction of parameters does not lead to a significant change in training time, as the duration of a backward pass is around 65-70 ms for a batch size of $64 \times 3 \times 224 \times 224$ on an NVidia GTX 1080 Ti. PFNet18 needs an additional time of 10 ms per batch. This is interesting because the number of mult-adds of PFNet18 is only $0.26 \cdot 10^9$ while ResNet18 has $1.81 \cdot 10^9$ mult-adds. Counting Mult-Add operations means counting the FLOPs of a model and to divide the result by 2. Although PFNet18 requires much fewer computations in total, its implementation requires more nodes in the computational graph and more distinct GPU calls. Our implementation seems to be not very efficient on current GPU hardware, which is optimized for a few but large tensor operations. Therefore, we assume that there is still much room for optimizations in the implementation of PFMs.

\subsection{Aliasing effects}

\begin{figure}[tb]
	\centering
	\adjustimage{trim=1cm 1cm 1cm 1cm, clip, width=0.6\linewidth}{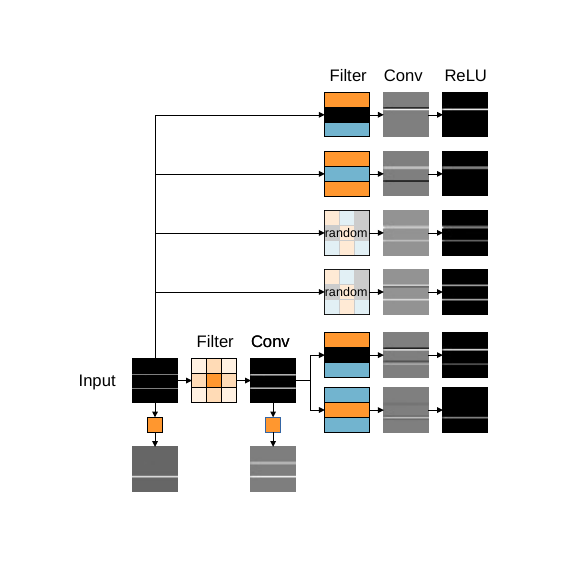}
	\caption{Aliasing effects. Various stride 2 convolutional operations are applied on the input image showing two horizontal lines. Orange color denotes positive kernel weights, black denotes zero, and blue negative weights.}
    \label{fig:Aliasing}
\end{figure}

Aliasing may occur in CNNs due to spatial sub-sampling. In PFNet18 and ResNet18 there are 3 skip connections with a $1 \times 1$ convolution and a stride of 2 where information is lost. The aliasing effects do not always seem to affect the classification performance. To test this, we blur the input data of these 3 convolution operations with a $3 \times 3$ Gaussian filter and train the models again. For ResNet18, there is no significant change in the test accuracy on the Flowers102 dataset. For PFNet18, however, there was a great improvement in test accuracy from $72.40 \pm 0.59$ to $80.66 \pm 0.35$.

The aliasing issue is studied in Figure \ref{fig:Aliasing} where different convolution operations with stride 2 are applied to an input image. The input image contains two white lines on a black background. The $1 \times 1$ convolution does only capture one of the lines because of spatial sampling. With stride 2, our $1 \times 3 \times 3$ pre-defined filters do not always capture both lines, either. This is a problem for PFNet18 because each input channel is convolved with exactly one $1 \times 3 \times 3$ pre-defined filter. The architecture lacks other convolution steps on the same input channel that could add redundancy. This means that information can be irrevocably lost. Thus, PFNet18 relies on the skip connections to compensate for aliasing effects and additional aliasing within the skip connections must be avoided.

Our results suggest that ResNet18 is more robust against aliasing in the skip connections. In Figure \ref{fig:Aliasing} we see that random $3 \times 3$ filters often capture both lines, which illustrates that ResNet18 has a lower risk of losing image information by aliasing. In addition, each convolution layer from ResNet18 convolves each input channel as many times as there are output channels. This gives more opportunities to propagate information from each input channel to the next layer.

\subsection{Ablation study}
\label{sec:results:ablation_study}

An ablation study is conducted to study the relevance of single elements in the PFM as well as the importance of the choice of pre-defined filters. Table \ref{tab:ablation_study} presents the results of the ablation study. The experiments are conducted on the Flowers102 dataset with the same hyperparameters as in the previous experiments. Similar to Table \ref{tab:datasets_results} the results show average values for five different seeds.

First, we show that the choice of pre-defined kernels matters. We pick 16 random pre-defined kernels from a uniform distribution. These kernels are frozen during training. Only the $1 \times 1$ convolutions are adjusted to the Flowers102 dataset. In this setting, the performance drops to $73.98\pm1.24\%$. However, interestingly, it is still as good as ResNet18.

In another experiment, the random pre-defined kernels are unfrozen, such that the model can determine the 16 kernels during training. Note that we give each PFM its own 16 kernels to optimize. This model gave a test accuracy of $74.54\pm0.88\%$, which is a small improvement. When the pre-defined filters are initialized with the edge filters and allowed to be optimized during training, the test accuracy is $80.11\pm0.62$, which is similar to the default experiment. Our findings indicate that the proposed set of filter kernels provides a beneficial bias for the Flowers102 dataset.

Second, we study the importance of the first ReLU in the PFMs. The first ReLU is removed from all PFMs. This means that the convolution with the pre-defined kernels and the subsequent $1 \times 1$ convolution form one linear operation. The resulting filter will have a kernel that is differently organized than the convolution kernels in ResNet18. The performance drops to $75.21\pm0.23$. The ReLU seems to be important, possibly due to the increased amount of non-linearity.

\begin{table}[tb]
\caption{Test accuracy of variants of PFNet18 and ResNet18 on the Flowers102 dataset for training from scratch.}
\label{tab:ablation_study}
\centering

\begin{tabular}{rlc}
& &\\
Model & Description & Accuracy \\
\midrule
PFNet18 & No aliasing, default & \textbf{80.66$\pm$0.35} \\
PFNet18 & Aliasing  & 72.40$\pm$0.59 \\
\midrule
PFNet18 & First ReLU removed, no aliasing  & 75.21$\pm$0.23 \\
\midrule
PFNet18 & 16 Trainable filters, edge init., no aliasing & 80.11$\pm$0.62 \\
PFNet18 & 16 Trainable filters, random init., no aliasing & 74.54$\pm$0.88 \\
PFNet18 & 16 frozen filters, random init., no aliasing & 73.98$\pm$1.24 \\

\midrule
ResNet18  & Default  & 73.40$\pm$0.34 \\
ResNet18  & No aliasing  & \textbf{74.36}$\pm$0.57 \\

\end{tabular}
\end{table}

\section{Discussion}
\label{sec:discussion}

Pre-defined filter kernels lead to significant performance improvements on half of the datasets that we evaluated. This improvement is achieved with only 16 different spatial $1 \times 3 \times 3$ filter kernels and only $13\%$ of the trainable weights of ResNet18. Despite the restriction of PFCNNs to only learn combinations of pre-defined filter outputs, discriminative features emerge during training. For PFCNNs, the ability to perform visual recognition is based on the appropriate combination of filter outputs. These findings provide a new perspective on the information processing within deep CNNs and they show once again, that many weights in conventional CNNs are redundant.

We found that aliasing can significantly reduce the performance of PFCNNs. Since this effect is much lower in CNNs, one can assume that the CNNs learn how to deal with aliasing; which, however, implies that resources need to be spent to deal with the problem.

Feature visualization shows that the combinations of the pre-defined filter outputs yield complex, object-specific features. Note that these features emerged from only 20 training images per class. Compared to ResNet18, our PFCNNs seem to also consider the shape of the recognized objects, while ResNet18 seems to focus on textures.

We discovered that the choice of pre-defined filters matters. When using random kernels instead of edge filters, the test accuracy on the Flowers102 dataset drops to $73.98\pm1.24\%$, which, however, is still as good as ResNet18, which is remarkable. We conclude that edge filters add a suitable bias to the image recognition problem.


\section{Conclusion}
\label{sec:conclusion}

We introduced a novel class of CNNs called Pre-defined Filter Convolutional Neural Networks (PFCNNs), which utilizes fixed pre-defined filters in all $n \times n$ convolution kernels with $n>1$ while keeping the end-to-end training paradigm. In our implementation, the PFNet18 architecture is a ResNet18 where we replaced the convolution operations with so-called Pre-defined Filter Modules (PFMs). These modules consist of a depthwise convolution with pre-defined weights that are not changed during training; only the weights of the subsequent $1 \times 1$ convolution are adjusted.
PFNet18 has only 13\% of the weights of ResNet18 but outperforms ResNet18 on the Caltech101, FGVC-Aircraft, and Flowers102 datasets with an absolute increase of $10\%$. On the CIFAR10, CUB-200-2011, and Stanford Cars datasets, PFNet18 does not reach higher performances than ResNet18 but still performs well with much fewer parameters. The results demonstrate that our choice of taking edge filters as pre-defined filter kernels is a useful bias for image data. Interestingly, the pre-defined edge filters are useful biases not only in the first but also in the higher layers of the CNN. Our results imply that it is unnecessary to train the spatial kernels of a CNN to reach reasonable test accuracies on image data, which saves most trainable weights. In contrast to pruning, where weights are eliminated, we save weights by excessive weight sharing using a small pool of 16 pre-defined kernels. Our approach uses only $13\%$ of the parameters of ResNet18, which may be useful for mobile devices and other applications where model size is critical.

Many questions regarding PFCNNs arise, e.g., the reduction of weights by the PFMs could be combined with more recent, efficient architectures to get very light models with much fewer parameters. These models are in stark contrast to the usual over-parameterized approaches but still achieve reasonable results. We hope that PFCNNs will lead to interesting comparative studies and a better understanding of the way information is processed internally by CNNs.
It is left to future research to explore how PFCNNs perform on large-scale datasets such as ImageNet \cite{russakovsky_imagenet_2015}, and how PFCNNs perform in transfer-learning scenarios. 
Another question involves how the choice (and number) of filters affect the performance of PFCNNs as well as the robustness against input perturbations and adversarial attacks.
In addition, the width of the networks could be increased to allow for more filters to be linearly combined in each layer. This might boost the performance of the PFCNNs according to Gavrikov and Keuper \cite{gavrikov_rethinking_2023} who found that increasing the width of networks with fixed, random spatial kernels leads to an improvement of the test accuracy on CIFAR10.
Also, the question remains to what extent the results can be transferred to other domains, for instance, audio processing.
Overall it seems that the power of deep networks mainly lies in their ability to learn how to combine filters, rather than in the learning of spatial convolution kernels.

\section*{Acknowledgment}
The work of Christoph Linse was supported by the Bundesministerium f{\"u}r Wirtschaft und Klimaschutz through the Mittelstand-Digital Zentrum Schleswig-Holstein Project.
The Version of Record of this contribution was presented at the 2023 International Joint Conference on Neural Networks (IJCNN),
and is available online at https://doi.org/10.1109/IJCNN54540.2023.10191449

\bibliographystyle{IEEEtran}
\bibliography{References}

\end{document}